\documentclass{article}

\usepackage{arxiv}

\usepackage[utf8]{inputenc} 
\usepackage[T1]{fontenc}    
\usepackage{hyperref}       
\usepackage{url}            
\usepackage{booktabs}       
\usepackage{amsfonts}       
\usepackage{nicefrac}       
\usepackage{microtype}      
\usepackage{lipsum}

\usepackage{caption,subcaption,amsmath,graphicx,url,times}
\usepackage{color}

\title{ADVERSARIAL ATTACKS IN SOUND EVENT CLASSIFICATION}

\author{
  Vinod Subramanian\thanks{This work has received funding from the European Union's Horizon 2020 research and innovation programme under the Marie Skłodowska-Curie grant agreement No. 765068. EB is supported by RAEng Research Fellowship RF/128.} \\
  Centre for Digital Music\\
  Queen Mary University of London\\
  UK\\
  \texttt{v.subramanian@qmul.ac.uk} \\
   \And
 Emmanouil Benetors\\
  Centre for Digital Music\\
  Queen Mary University of London\\
  UK\\
  \texttt{emmanouil.benetos@qmul.ac.uk} \\
   \AND
   Ning Xu \\
   ROLI Ltd. \\
   UK\\
   \texttt{ning@roli.com} \\
   \And
   SKoT McDonald \\
   ROLI Ltd. \\
   UK\\
   \texttt{skot.mcdonald@roli.com } \\
   \And
   Mark Sandler \\
   Queen Mary University of London \\
   UK \\
   \texttt{mark.sandler@qmul.ac.uk} \\
}

\begin{document}
\maketitle

\begin{abstract}
Adversarial attacks refer to a set of methods that perturb the input to a classification model in order to fool the classifier. In this paper we apply different gradient based adversarial attack algorithms on five deep learning models trained for sound event classification. Four of the models use mel-spectrogram input and one model uses raw audio input. The models represent standard architectures such as convolutional, recurrent and dense networks. The dataset used for training is the FSDKaggle2018
released for task 2 of the DCASE 2018 challenge and the models used are from participants of the challenge who open sourced their code. Our experiments show that adversarial attacks can be generated with high confidence and low perturbation. In addition, we show that the adversarial attacks are very effective across the different models.
\end{abstract}


\section{Introduction}
Sound event classification has applications in surveillance to detect gunfire or explosions, in self driving cars to detect car horns, in curating large music libraries for ease of access, for censoring harmful content in videos etc. In recent years, large datasets such as Urban Sound 8k \cite{Salamon:UrbanSound:ACMMM:14}, FSDKaggle2018 \cite{Fonseca}, YouTube-8M \cite{DBLP:journals/corr/Abu-El-HaijaKLN16} and Audio set \cite{gemmeke2017audio} encouraged research in sound event classification and deep learning is the most popular approach to solving the task.

We need to ensure that the models used for these tasks are robust and secure because mistakes can be costly. In terms of security risks, a surveillance system cannot afford to classify an explosion as a benign sound, a self driving car cannot ignore a car horn etc. In terms of robustness, a poorly curated sound library is difficult for a user to navigate while an algorithm that is not able to detect harmful content in videos forces reliance on human labour which is time and cost intensive. Currently, we use different evaluation metrics such as accuracy, precision, recall and F-score to analyze the classification performance of models. However, these metrics do not provide insight into the underlying features learned by different layers of the models.

Recent research has shown that deep learning models can be fooled by perturbations of the input, this is called an adversarial attack \cite{szegedy_intriguing_2013,goodfellow_explaining_2014}. Given the popularity of deep learning approaches for sound event classification \cite{virtanen2018computational}, it is important to investigate the effect of adversarial attacks on models which classify sound.

\subsection{Sound Event Classification}
Large datasets such as ImageNet \cite{imagenet_cvpr09} helped deep learning cement itself as an effective technique to solve image classification tasks. Following that work, research has shown that convolutional neural networks are effective at audio classification tasks on large datasets as well \cite{hershey_cnn_2017,salamon_deep_2017}. 

The best submissions of task 2 of DCASE 2018 challenge follow the trend of convolutional neural networks for audio classification \cite{Iqbal2018GeneralpurposeAT,jeong2018audio}. They experiment with more sophisticated convolutional architectures such as convolutional recurrent networks \cite{liang2015recurrent}, dense networks \cite{huang2017densely} and gated convolutions \cite{dauphin2017language}. Most of the proposed models use logmel spectrograms as the input and use mixup augmentation \cite{zhang2017mixup} while training.

\subsection{Adversarial Attacks}
Szegedy et al. \cite{szegedy_intriguing_2013} discovered that in image recognition tasks by applying an imperceptible non-random perturbation to the input the output of the network can be changed. The term ``Adversarial examples” is used to describe these perturbed inputs. The explanation for adversarial examples was that they were caused by blind-spots while training these deep learning models. They introduced the L-BFGS algorithm which is an approximation of the Broyden–Fletcher–Goldfarb–Shanno (BFGS) algorithm \cite{fletcher2013practical} to generate adversarial examples. Goodfellow et al. \cite{goodfellow_explaining_2014} introduce a family of fast gradient methods to linearly perturb a deep learning model to generate adversarial examples and showed that adversarial examples are a result of linearity in high dimensions.

Following these initial findings, more sophisticated attack algorithms were introduced. The saliency map method \cite{papernot_limitations_2015} is based on saliency maps \cite{simonyan_deep_2013} which computes a score for each point in the input to the output. The goal is to change the input so that the score of the current label decreases and the score of the desired label is maximized. Deepfool is an attack where the classifier is iteratively linearized to generate a perturbation to minimize the confidence of the current label until the classifier is fooled \cite{moosavi2016deepfool}. The Carlini and Wagner attack \cite{carlini_towards_2016} uses the Adam optimization algorithm \cite{kingma2014adam} with an $L2$ distance term to reduce distortion and it iteratively updates the perturbation until the classifier is fooled. All of these adversarial attacks were extensively tested for image classification on models trained on the MNIST \cite{lecun_mnist}, CIFAR \cite{krizhevsky2009learning} and ImageNet \cite{imagenet_cvpr09} datasets; however, very little research has explored the effects of adversarial attacks in sound event classification.

In the audio domain Kereliuk et al. \cite{kereliuk_deep_2015} implemented adversarial attacks for music genre classification. They modified the L-BFGS \cite{szegedy_intriguing_2013} and fast gradient algorithms \cite{goodfellow_explaining_2014} to generate perturbations on the logmel spectrogram input. By ensuring that the logmel spectrogram generated is invertible into audio using the Griffin-Lim algorithm \cite{griffin1984signal}, they were able to create adversarial examples for audio. Du et al. \cite{du_siren} introduced the SirenAttack that uses Particle Swarm Optimization \cite{eberhart1995new}, which is a heuristic and stochastic algorithm, to generate adversarial examples. They test the effectiveness of the attack on speech recognition, sound event classification and music genre classification.

In this paper, we focus on the FSDKaggle2018 \cite{Fonseca} dataset, released for task 2 of the DCASE 2018 challenge\footnote{http://dcase.community/challenge2018/task-general-purpose-audio-tagging-results}, titled ``General-purpose audio tagging of Freesound content with AudioSet labels". We chose this task because the abundance of high accuracy models that were open sourced allowed us to run the adversarial attack algorithms against more models, which gave us more data points for our experiments. The main contributions of this paper is to perform an analysis of four different adversarial attack algorithms on five different models trained on the same dataset. We show that across all models we can create adversarial examples with a very high signal-to-noise ratio (SNR). In addition, we observe that a small percentage of the adversarial attacks are transferable between the different models.

\section{Methodology}
\subsection{Adversarial attack algorithms}
Let us denote classifier for a single label task as $D$ which maps an $m$ dimensional audio input $x \in [-1,1]^m$ to an output with $k$ classes $y \in \{1,2,\dots,k\}$ through the equation $y = D(x)$. Then, the goal of an adversarial attack is to create a perturbation $r$ such that the new input is given by $x_{adversarial} = x + r$ then $y_{new} = D(x_{adversarial})$, where  $y_{new} \neq y$. Ideally the perturbation $r$ is imperceptible. There are different typs of adversarial attacks. They can be classified based on the the optimization goal as:
\begin{enumerate}
    \item Untargeted attack: The goal is to reduce the confidence of the current label until the classifier is fooled into a new label.
    \item Targeted attack: In targeted attack we pick a desired output and the goal is to maximize the confidence of the desired output. The attack is successful when the classifier changes the output label to the desired label.
\end{enumerate}
Adversarial attacks can be classified based on the knowledge of the model they are attacking as:
\begin{enumerate}
    \item Perfect knowledge: The adversarial attack has complete knowledge of model architecture and weights of each layer.
    \item Limited knowledge: The adversarial attack knows the model architecture and outputs only.
    \item Zero knowledge: The adversarial attack has no knolwedge of the model architecture or output.
\end{enumerate}

In this paper we perform untargeted attacks using white noise, fast gradient sign method (FGSM) \cite{goodfellow_explaining_2014}, Deepfool \cite{moosavi2016deepfool} and Carlini \& Wagner (C\&W) \cite{carlini_towards_2016} attack. The white noise and fast gradient method serve as baselines to compare the Deepfool and Carlini \& Wagner  attack. The targeted attacks are performed using the L-BFGS and Carlini \& Wagner methods. Both the targeted and untargeted attacks are performed with perfect knowledge of the models. Finally, we test the adversarial examples generated for one model on the other models which is a zero knowledge attack. 

\subsection{Model architectures}
We use the models proposed by Iqbal et al. \cite{Iqbal2018GeneralpurposeAT} and Jeong and Lim \cite{jeong2018audio} for task 2 of DCASE 2018 challenge.  Table \ref{table:perf} shows the training and testing accuracy of the models described below.
\begin{enumerate}
    \item Iqbal et al: We use three of the proposed architectures by Iqbal et al. \cite{Iqbal2018GeneralpurposeAT}, the VGG13, CRNN and GCNN architectures. VGG13 is a convolutional neural network inspired by the network proposed for image classification \cite{simonyan2014very} of the same name. CRNN is a convolutional recurrent neural network where the convolutional layers are followed by a bidirectional recurrent layer. GCNN is a gated convolutional neural network which is a modification of the convolutional layer inspired by the gated layer in Long short-term memory (LSTMs) \cite{hochreiter1997long}.
    
    The input features to the models start with resampling the audio to 32kHz and computing the spectrogram where the spectrogram has a window size of 1024 samples and a hop size of 512 samples. The spectrogram is converted into a mel spectrogram with 64 mel bands. Finally, the logmel spectrogram is computed and normalized by the absolute maximum value of the logmel spectrogram.

    \item Jeong and Lim: They propose a modification of the DenseNet architecture \cite{jeong2018audio}. The DenseNet module concatentates the input to the output in order to stabilize the training process as the model gets deeper. The output is generated through a multi-head classifier where the softmax values from the 8 heads are averaged to give the final prediction. 
    
    Two configurations of the model are used, one with raw audio input (dense-wav) and the other with logmel spectrogram input (dense-mel). In both configurations the audio is resampled to 32kHz and in the case of the logmel spectrogram configuration batch normalization is applied before the logmel spectrogram is computed.
    
    The DenseNet architectures do not have a single logits layer because of the multi-head classifier so we use the averaged output probabilites of the classifier for the adversarial attacks instead.
\end{enumerate}
\begin{table}[h]
\centering
\begin{tabular}{|l|l|l|} 
\hline
\textbf{Model}    & \textbf{Training} & \textbf{Test}  \\ 
\hline
\textbf{VGG13}         &     0.9714              &  0.8093                 \\ 
\hline
\textbf{CRNN}          &       0.9768           &          0.8437         \\ 
\hline
\textbf{GCNN}          &         0.9803          &          0.8437         \\ 
\hline
\textbf{dense\_mel} &       0.9876            &        0.89875           \\ 
\hline
\textbf{dense\_wav} &        0.9698           &     0.86125         \\
\hline
\end{tabular}
\caption{Model performance on training and test data}
\label{table:perf}
\end{table}

\section{Experiments}
\subsection{Experiment 1: Untargeted attacks}
We perform untargeted attacks using white noise, Fast Gradient Sign Method, Deepfool and Carlini and Wagner. There are 41 classes in the FSDKaggle2018 dataset, in this experiment we use 6 randomly chosen examples per class for a total of 246 audio files. We use a subset because it would be too time intensive to run adversarial attacks for the whole dataset. The goal of the untargeted attack experiment is to investigate if there are any classes that get frequently confused with each other and if we can identify any intuitive pattern in the way the new labels are created.
\subsection{Experiment 2: Targeted attacks}
The targeted attacks are performed using the L-BFGS and Carlini \& Wagner methods. We use a subset of the untargeted attack dataset by picking 6 of the 41 classes for a total of 36 audio files. In the targeted attack for an audio file of a particular class we run the adversarial attack 5 times for each of the other 5 classes. The 6 classes picked is a music subset of Bass drum, Snare drum, Cello, Violin, Clarinet and Oboe. The reason for picking these instruments is that we want to investigate if the properties of the adversarial examples can be acoustically explained. For example, is the SNR to turn Cello into Violin higher than to turn it into a Snare drum because the Cello and Violin sound more similar.
\subsection{Experiment 3: Transferability of untargeted attacks}
So far we have generated adversarial attacks that are unique to each classifier. In this experiment we take the attacks generated from Deepfool and Carlini \& Wagner of one model and test them on the other four models. The goal of the experiment is to shed light on the cause of adversarial examples. If adversarial examples are not transferrable at all then that means that they are unique to a particular deep learning architecture. However, if they are transferable that would mean there is some underlying connection between the models which could be attributed to input pipeline, specific layers or even the dataset. 
\section{Results and Discussion}
\begin{table}[h]
\centering
\begin{tabular}{|l|l|l|l|l|} 
\hline
\textbf{Attack} & \textbf{Success \%}  & \textbf{New} & \textbf{GT New} &\textbf{SNR}  \\ 
\hline
\textbf{White noise}    & 21.14      &    0.3709     &  0.1581 & 20.00\\ 
\hline
\textbf{FGSM}     & 31.14      & 0.3073         &  0.0984     & 19.99      \\ 
\hline
\textbf{Deepfool}     & 99.99           & 0.3502         &  0.2476   &41.81       \\ 
\hline
\textbf{C\&W}     &99.51        &0.7103          & 0.0189    &46.23      \\
\hline

\end{tabular}
\caption{The success rate, new label confidence (New), original label confidence on new input (GT New) and SNR across the five models for each attack.}
\label{table:adv}
\end{table}

\begin{table}[h]
    \begin{subtable}[h]{0.45\textwidth}
        \centering
        \begin{tabular}{|l|l|l|l|l|l|} 
        \hline
        \textbf{Model} & \textbf{Acc.}        & \textbf{GT} & \textbf{New} & \textbf{GT new} &\textbf{SNR}  \\ 
        \hline
        \textbf{VGG13}    & 100    &  0.6852      &    0.2190      &  0.1681 & 52.47\\ 
\hline
\textbf{CRNN}     & 100           &  0.7434      & 0.2913         &  0.2181     & 49.40      \\ 
\hline
\textbf{GCNN}     & 100       &  0.7173      & 0.2340         &  0.1750   &50.97        \\ 
\hline
\textbf{dense-mel}     &99.6  & 0.9522       &0.4982          & 0.3628    &55.42        \\
\hline
\textbf{dense-wav}     &100  &0.9426        & 0.5180         &0.3148 &41.81          \\
\hline
\end{tabular}
       \caption{Deepfool}
       \label{table:un-deep}
    \end{subtable}
    \hfill
    \begin{subtable}[h]{0.45\textwidth}
        \centering
        \begin{tabular}{|l|l|l|l|l|l|} 
\hline
\textbf{Model} &\textbf{ Acc.}        & \textbf{GT} & \textbf{New} & \textbf{GT new} &\textbf{SNR}  \\ 
\hline
\textbf{VGG13}    & 100    &  0.6852      &    0.5876      &  0.0208 & 52.10\\ 
\hline
\textbf{CRNN}     & 100           &  0.7434      & 0.6026         &  0.0213     & 50.64      \\ 
\hline
\textbf{GCNN}     & 100       &  0.7173      & 0.5908         &  0.0207   &51.32        \\ 
\hline
\textbf{dense-mel}     &99.60  & 0.9523       &0.9216          & 0.0119    &55.67        \\
\hline
\textbf{dense-wav}     &97.97  &0.9426        & 0.8932         &0.0199 &45.62          \\
\hline
\end{tabular}
        \caption{Carlini \& Wagner}
        \label{table:un-car}
     \end{subtable}
     \caption{Analysis of attack performance across each model}
     \label{table:un}
\end{table}

\begin{figure}[t!]
    \centering
    \begin{subfigure}[t]{0.24\textwidth}
       \centering
        \centerline{\includegraphics[width=\columnwidth,trim={0 0 10cm 21cm},clip]{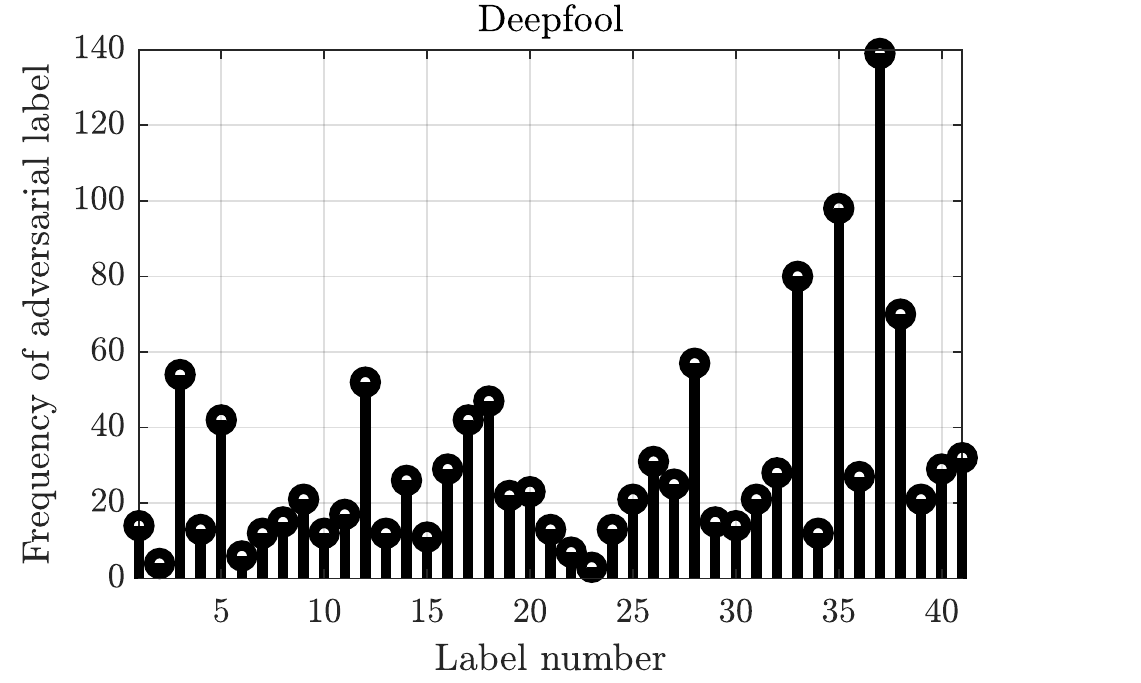}}
    \end{subfigure}%
    ~
    \begin{subfigure}[t]{0.24\textwidth}
        \centering
        \centerline{\includegraphics[width=\columnwidth,trim={0 0 10cm 21cm},clip]{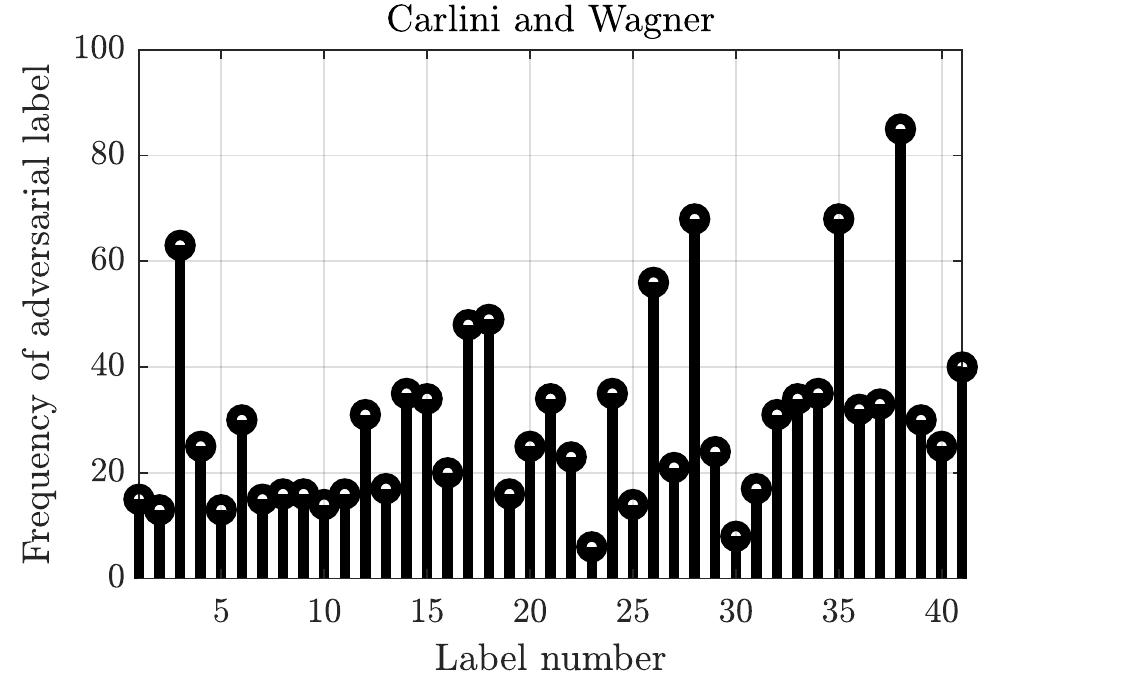}}
    \end{subfigure}
    \caption{Label distribution of the adversarial examples}
    \label{fig:lab-dist}
\end{figure}

The results for experiment 1 are displayed in tables \ref{table:adv} and \ref{table:un} and figure \ref{fig:lab-dist}. 

Table \ref{table:adv} provides a summary of performances of the untargeted adversarial attack algorithms. It compares the success rate, new label confidence, the ground truth label confidence in the adversarial audio and the perturbation measured by SNR. The table shows that a simple algorithm such as the fast gradient method can outperform white noise at the same SNR and that more sophisticated algorithms are almost a 100\% successful and less noisy. This is an indicator of the fact that it is easy to generate adversarial examples in audio that are almost indistinguishable to the human ear compared to the original. Select examples can be heard here \footnote{Link will be added}.

With Deepfool and Carlini \& Wagner attacks we have the performance compared across models in table \ref{table:un} where GT, New and GT new refer to the average output probability of the label of original audio, label of adversarial audio and ground truth label on the adversarial audio. The SNR is lowest for the raw audio configuration of the DenseNet and the highest for the logmel configuration of the DenseNet. This indicates that raw audio models might be more robust to adversarial attacks than logmel spectrogram models. 

In figure \ref{table:un-deep} and \ref{table:un-car} we plot the label numbers of the different adversarial examples generated from the untargeted attacks. The Deepfool attack has a skewed distribution with a large peak at label 37 which corresponds to the label ``Tearing". The Carlini and Wagner attack has a more even distribution of labels with a peak at 38 which corresponds to ``Telephone". There are no obvious pairs of audio that get confused with each other. This suggests that there is no acoustically explainable reason for how the labels are changing.

\begin{figure}[h!]
  \centering
  \centerline{\includegraphics[width=0.5\columnwidth,trim={0 0 10cm 21cm},clip]{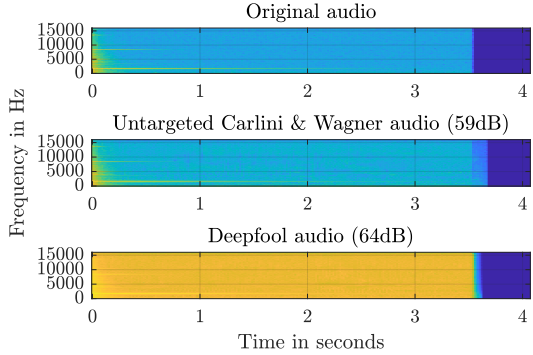}}
  \caption{This is the spectrogram of a Glockenspiel and its two adversarial audio files generated using the Carlini \& Wagner attack and Deepfool attack. The SNR in dB of the adversarial attack wrt the original audio is indicated in brackets}
  \label{fig:spec}
\end{figure}

Figure \ref{fig:spec} shows an example of the spectrogram of the original audio of a Glockenspiel and the untargeted Carlini \& Wagner and Deepfool attack. In this example we observe that the Carlini \& Wagner attack adds noise to the silence at the end and spreads the noise thin across the audio file making the spectrogram look similar to the original. Thee Deepfool does not add much noise to the silence and instead spreads it across the spectrum.

\begin{table}[h]
    \begin{subtable}[h]{0.45\textwidth}
        \centering
        \resizebox{\columnwidth}{!}{
        \begin{tabular}{|l|l|l|l|l|l|l|} 
\hline
           & \textbf{Bass d} & \textbf{Cello} & \textbf{Clarinet} & \textbf{Oboe} & \textbf{Snare} & \textbf{Violin}  \\ 
\hline
\textbf{Bass d}  &   NA      &  109.0    & 104.9  &105.9     & 102.3    &105.9\\ 
\hline
\textbf{Cello}&   105.8  & NA         & 108.4&  105.9& 107.4  & 106.6 \\ 
\hline
\textbf{Clarinet}&    108.3 &    108.9  & NA    & 106.5 & 109.1 & 107.6 \\ 
\hline
\textbf{Oboe}     &    100.2 &    102.4  & 102.3& NA     & 100.6 &104.8\\ 
\hline
\textbf{Snare}       &    93.2  &    94.4   &93.0  & 95.4  & NA   & 95.1     \\ 
\hline
\textbf{Violin}   &    97.2  &    100.5  &98.8  & 101.1 &98.2 & NA        \\
\hline
\end{tabular}
}
       \caption{Carlini \& Wagner}
       \label{table:un-deep}
    \end{subtable}
    \hfill
    \begin{subtable}[h]{0.45\textwidth}
        \centering
        \resizebox{\columnwidth}{!}{
        \begin{tabular}{|l|l|l|l|l|l|l|} 
\hline
           & \textbf{Bass d} & \textbf{Cello} & \textbf{Clarinet} & \textbf{Oboe} & \textbf{Snare} & \textbf{Violin}  \\ 
\hline
\textbf{Bass d}  &   NA      &  60.8   & 59.7  &59.9     & 60.9    &61.7\\ 
\hline
\textbf{Cello}&   55.4  & NA         & 55.8&  54.3& 56.0  & 57.0 \\ 
\hline
\textbf{Clarinet}      &    57.3 &    56.7  & NA    & 57.6 & 57.8 & 58.1 \\ 
\hline
\textbf{Oboe}     &    54.5 &    54.8 & 55.4& NA     & 54.6 &55.4\\ 
\hline
\textbf{Snare}       &    52.1  &    52.2   &50.3  & 50.1  & NA   & 51.7     \\ 
\hline
\textbf{Violin}   &    56.0  &    58.6  &57.8  & 57.6 &56.7 & NA        \\
\hline
\end{tabular}
}
        \caption{L-BFGS}
        \label{table:un-car}
     \end{subtable}
     \caption{Average SNR to perform targeted attack between a pair of labels across all models}
     \label{table:tarcm}
\end{table}

\begin{figure}[t!]
    \centering
    \begin{subfigure}[t]{0.45\textwidth}
       \centering
        \centerline{\includegraphics[width=\columnwidth,trim={0 0 10cm 21cm},clip]{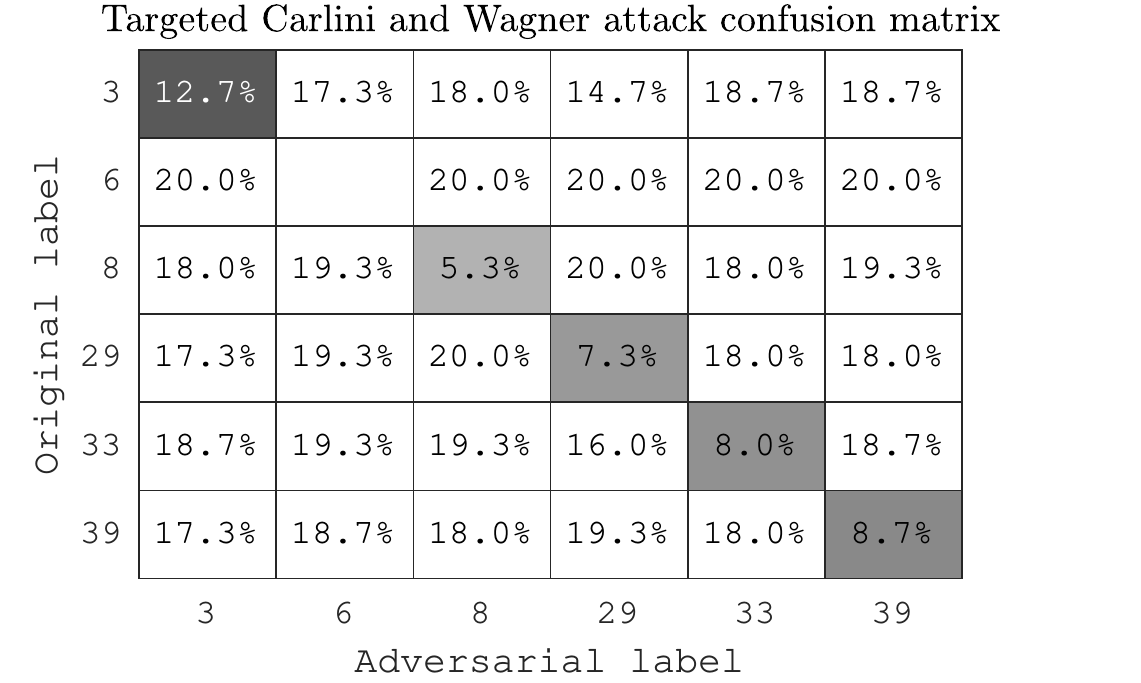}}
    \end{subfigure}%
    \hfill
    \begin{subfigure}[t]{0.45\textwidth}
        \centering
        \centerline{\includegraphics[width=\columnwidth,trim={0 0 10cm 21cm},clip]{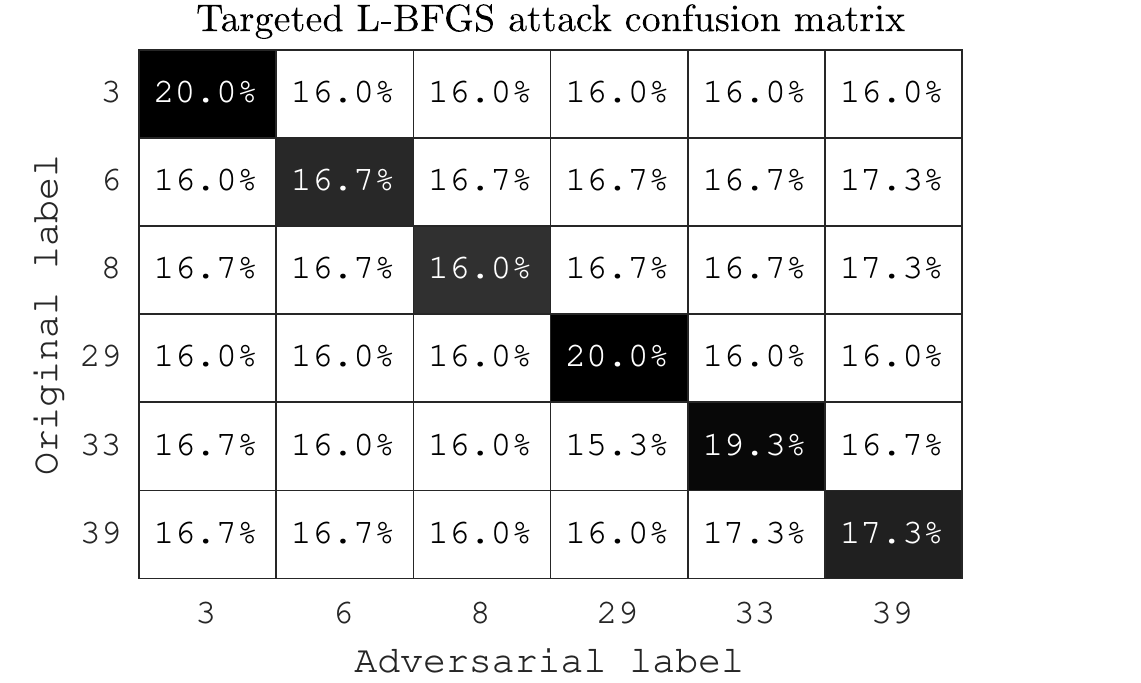}}
    \end{subfigure}
    \caption{Targeted attack success plot. 3: Bass drum, 6: Cello, 8: Clarinet, 29: Oboe,33: Snare drum, 39: Violin}
    \label{fig:tarcm}
\end{figure}

The results for experiment 2 are shown in in tables \ref{table:tarcm} and in figure \ref{fig:tarcm}.
Table \ref{table:tarcm} shows the average SNR to generate a targeted adversarial attack averaged across all models for every pair of labels. The Carlini \& Wagner attack requires lesser perturbation to generate the attacks than L-BFGS. Between the two models we observe that the Snare drum requires the most perturbation to fool into the other classes. We also observe that acoustic similarity between instruments does not translate into requiring lesser perturbation which further supports the results from experiment 1 that the adversarial labels are not acoustically explainable. 

Figure \ref{fig:tarcm} show the confusion matrix of the targeted attack. The Carlini \& Wagner attack has a higher success rate than the L-BFGS algorithm. We observe that the success rate is much lower for the targeted attacks compared to the Carlini \& Wagner and Deepfool untargeted attacks. Targeted attacks are typically harder to generate because of the stricter constraints on the output of the model.

\begin{table}[h!]
\centering
\resizebox{0.5\columnwidth}{!}{
\begin{tabular}{|l|l|l|l|l|l|} 
\hline
 & \textbf{VGG13}         & \textbf{CRNN} & \textbf{GCNN} & \textbf{dense-mel} &\textbf{dense-wav}  \\ 
\hline
\textbf{VGG13}  &NA  & 5.69    &  9.96      &    1.02      &  3.05\\ 
\hline
\textbf{CRNN}     & 5.89           &  NA      & 6.71         &  0.81     & 2.03      \\ 
\hline
\textbf{GCNN}     & 6.30       &  4.88      & NA         &  1.63   &2.85        \\ 
\hline
\textbf{dense-mel}     &0.20  & 0.41       &0.41          & NA   &0.61        \\
\hline
\textbf{dense-wav}     &0  &0        & 0         &0.20 &NA          \\
\hline
\end{tabular}
}
\caption{Percentage of adversarial attacks transferable between models for the untargeted Deepfool and Carlini \& Wagner attacks. }
\label{table:trans}
\end{table}

Experiment 3 results are shown in in table \ref{table:trans}. It indicates the success rate of the untargeted Carlini \& Wagner and Deepfool attacks of each model applied on the remaining models. The y-axis is the model on which the attack is being applied and the x-axis is the model whose adversarial examples are being used. We observe that the transfer percentage is not very high. However, the fact that any of the adversaries are transferable at all indicates that there is some underlying connection to the different models. The fact that the logmel models are most unsuccessful on the raw audio model suggests that similarities in the input representation contribute to transferability and the fact that the only transferable examples to the raw audio DenseNet is also a DenseNet model indicates that model architecture similarities contribute to transferability.

\section{Conclusion}
In this paper we have demonstrated the presence of adversarial attacks in sound event classification. We have shown that it is possible to generate these adversarial examples with very small perturbations and that we can change the label of an input to the desired input with high confidence. Further we have shown that there is no obvious acoustic explanation for these adversarial examples. This work has only begun to scratch the surface of the threat that adversarial attacks poses to sound event classification models.

In future work we want to focus on defenses against these attacks. We want to further investigate different input representations such as spectrogram and constant-Q transform and make concrete conclusions about how different input representations affect robustness of deep learning models. We will also look at how resilient the adversarial examples are to typical audio compression techniques at different bit-rates.

We hope this research encourages more work into investigating the robustness of deep learning models for sound event classification as well as a broader discussion about the responsible use of deep learning models for real world applications.

\bibliographystyle{unsrt}  
\bibliography{adversarial-attacks.bib}  


\end{document}